\pdfoutput=1

\documentclass[sigconf]{acmart}

\usepackage{amsmath,amsfonts,bm}

\def\eqref#1{equation~\ref{#1}}

\def\1{\bm{1}}

\def\mA{{\bm{A}}}
\def\mB{{\bm{B}}}

\def\mD{{\bm{D}}}

\def\mK{{\bm{K}}}

\def\mO{{\bm{O}}}
\def\mP{{\bm{P}}}
\def\mQ{{\bm{Q}}}
\def\mR{{\bm{R}}}
\def\mS{{\bm{S}}}

\def\mV{{\bm{V}}}
\def\mW{{\bm{W}}}
\def\mX{{\bm{X}}}

\def\mZ{{\bm{Z}}}

\DeclareMathAlphabet{\mathsfit}{\encodingdefault}{\sfdefault}{m}{sl}
\SetMathAlphabet{\mathsfit}{bold}{\encodingdefault}{\sfdefault}{bx}{n}

\newcommand{\R}{\mathbb{R}}

\usepackage{todonotes}
\usepackage{hyperref}
\usepackage{url}
\usepackage{multirow} 
\usepackage{xcolor} 
\usepackage{colortbl} 
\usepackage{tablefootnote}
\usepackage{graphicx}
\usepackage{soul}
\usepackage{float}
\usepackage{wrapfig}
\usepackage{lipsum}  
\usepackage{siunitx}
\usepackage{enumitem}
\usepackage{booktabs} 
\usepackage{array}    
\usepackage{xspace}
\definecolor{myyellow}{RGB}{255, 102, 102}
\definecolor{darkred}{rgb}{0.55, 0.0, 0.0}

\newcommand{\model}{TimelyGPT\xspace}
\AtBeginDocument{%
  }

\setcopyright{acmlicensed}
\copyrightyear{2024}
\acmYear{2024}
\acmDOI{XXXXXXX.XXXXXXX}

\acmConference[ACM BCB '24]{The 15th ACM Conference on Bioinformatics, Computational Biology, and Health Informatics}{Nov. 22-25, 2024}{Shenzhen, Guangdong, PR China}

\acmISBN{978-1-4503-XXXX-X/18/06}

\begin{document}

\title[TimelyGPT]{TimelyGPT: Extrapolatable Transformer Pre-training for Long-term Time-Series Forecasting in Healthcare}

\author{Ziyang Song, Qincheng Lu, Hao Xu, He Zhu}
\affiliation{%
  \institution{School of Computer Science, McGill University}
  \city{Montreal}
  \state{QC}
  \country{Canada}
  \postcode{H3A 2A7}
}

\author{David Buckeridge}
\affiliation{
  \institution{School of Population and Global Health, McGill University}
  \streetaddress{1140 Pine Avenue West}
  \city{Montreal}
  \state{Quebec}
  \country{Canada}
  \postcode{H3A1A3}
}

\author{Yue Li}
\affiliation{%
  \institution{School of Computer Science, McGill University}
  \institution{Mila Quebec AI institute}
  \city{Montreal}
  \state{QC}
  \country{Canada}
  \postcode{H3A 2A7}
}

\renewcommand{\shortauthors}{Song et al.}

\begin{abstract}
\textbf{Motivation:} Large-scale pre-trained models (PTMs) such as BERT and GPT have recently achieved great success in Natural Language Processing and Computer Vision domains. However, the development of PTMs on healthcare time-series data is lagging behind. This underscores the limitations of the existing transformer-based architectures, particularly their scalability to handle large-scale time series and ability to capture long-term temporal dependencies.\\
\textbf{Methods:} In this study, we present Timely Generative Pre-trained Transformer (\model). \model employs an extrapolatable position (xPos) embedding to encode trend and periodic patterns into time-series representations. It also integrates recurrent attention and temporal convolution modules to effectively capture global-local temporal dependencies. \\
\textbf{Materials:} We evaluated \model on two large-scale healthcare time series datasets corresponding to continuous biosignals and irregularly-sampled time series, respectively: (1) the Sleep EDF dataset consisting of over 1.2 billion timesteps collected from 197 whole-night polysomnographic sleep recordings, containing EEG, EOG, EMG, and event marker; (2) the longitudinal healthcare administrative database PopHR, comprising 489,000 patients randomly sampled from the Montreal population. \\
\textbf{Results:} Our experiments show that  during pre-training, \model excels in learning time-series representations from continuously monitored biosignals and irregularly-sampled time series data commonly observed in longitudinal electronic health records (EHRs), which can aid in healthcare time-series forecasting tasks. In  forecasting continuous biosignals, \model achieves accurate extrapolation up to 6,000 timesteps of body temperature during the sleep stage transition, given a short look-up window (i.e., prompt) containing only 2,000 timesteps. For irregularly-sampled time series, \model with a proposed time-specific inference demonstrates high top recall scores in predicting future diagnoses using early diagnostic records, effectively handling irregular intervals between clinical records. Together, we envision \model to be useful in a broad spectrum of health domains, including long-term patient health state forecasting and patient risk trajectory prediction.
\end{abstract}




\maketitle

\section{Introduction}
Time-series forecasting holds significant importance in healthcare, given its potential to trace patient health trajectories and predict medical diagnoses \cite{TS_PTM_survey, TS-TCC}. In the field of healthcare, there are two primary categories: continuously monitored and irregularly-sampled time series data. Continuous time-series, such as biosignals, have been extensively studied in various applications, including health monitoring \cite{biosignal_forecasting}, disease classification \cite{XSleepNet}, and physical activity prediction \cite{DeepPPG}. Irregularly-sampled time series are commonly found in clinical records, where spontaneous updates are made due to outpatient hospital visits or inpatient hospital stays \cite{RAINDROP}. The key challenge is to extract meaningful contextualized representations from these time-series to make accurate long-term forecasting. A promising approach is to adopt transfer learning \cite{TS_PTM_survey}. Initially, a model is pre-trained on large-scale datasets to learn contextualized temporal representations. This pre-trained model (PTM) is then fine-tuned to forecast target sequences.


\begin{figure*}[ht]
\centering
\includegraphics[width=\textwidth]{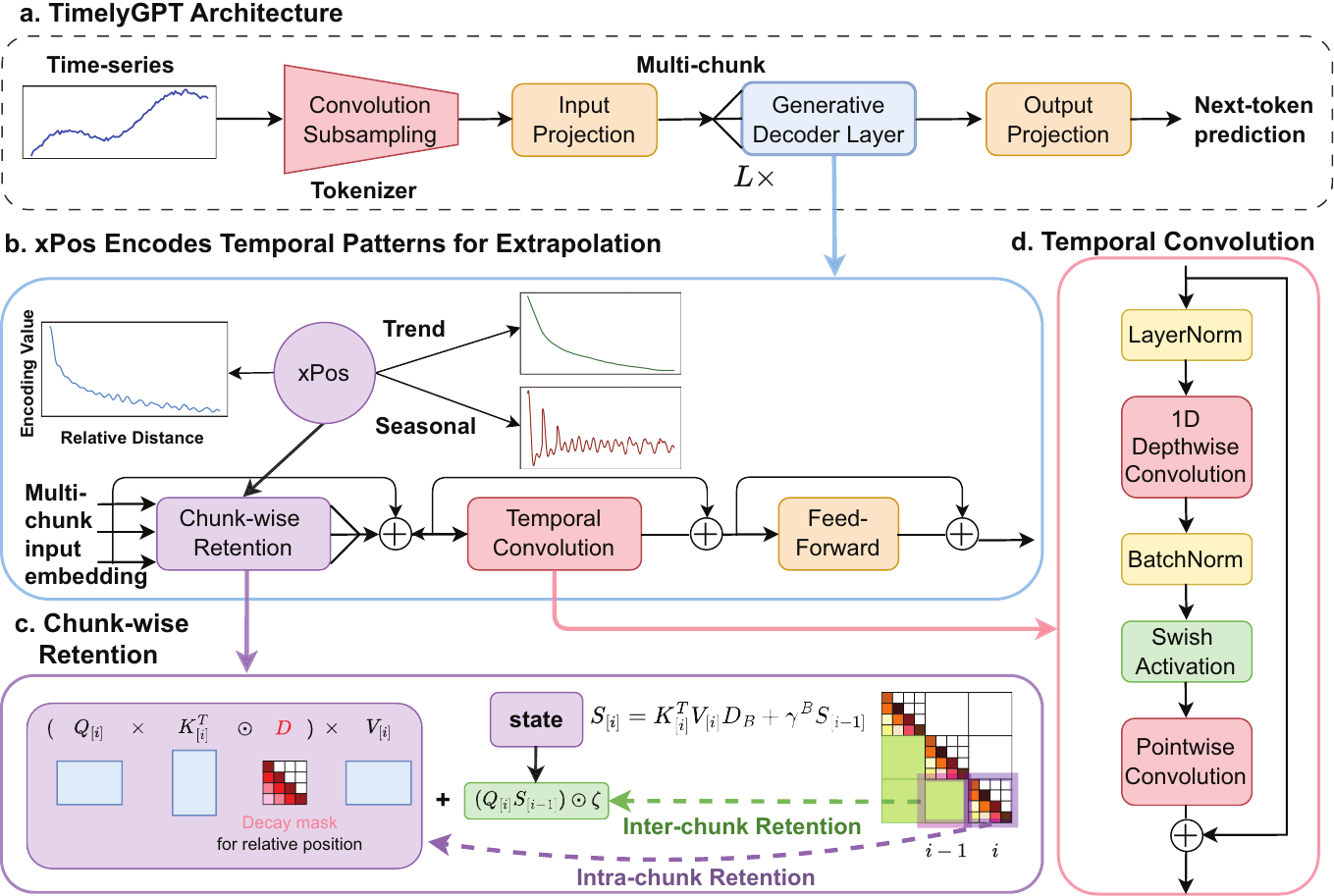}
\caption{\model overview. \textbf{a.} \model architecture. \model consists of a convolution-subsampling tokenizer followed by $L$ decoder layers, with detailed overflow provided in Appendix \ref{sec: overflow}. \textbf{b.} Generative decoder with xPos embedding. Each decoder layer is coupled with extrapolatable position embedding (Section \ref{sec:xPos}) that encodes  trend and periodic patterns into representations, facilitating forecasting with extrapolation ability. \textbf{c.} \textbf{Chunk-wise Retention}. This module consists of parallel intra-chunk Retention and recurrent inter-chunk Retention, effectively handling long sequences in continuously monitored biosignals (Appendix \ref{equivalence}). \textbf{d.} \textbf{Temporal Convolution} (Section \ref{sec: convolution}) captures nuanced local interactions from time-series representations.
}
\label{intro_figure}
\end{figure*}
 
The recent impressive achievements of Transformer PTMs in Natural Language Processing (NLP) and Computer Vision (CV) domains have inspired growing interest in time-series Transformer-based PTMs. Time-Series Transformer (TST) uses a mask-and-reconstruction pre-training strategy to extract contextualized representations from time series \cite{TST}. Cross-Reconstruction Transformer (CRT) learns temporal representations by dropping and reconstructing certain segments from time series \cite{CRT}. Additionally, Transformer PTMs have been applied to traffic \cite{traffic1}, tabular \cite{tabular1}, and speech time-series \cite{speech1, speech2}.

Transfer learning by pre-training on large time-series data followed by fine-tuning for long-term time series forecasting (LTSF) tasks is a promising avenue. However, existing studies primarily focus on training from scratch on limited data for LTSF tasks \cite{TS_PTM_survey}. These studies often introduce tailored architectures and attention modules to extract complex temporal dependencies \cite{informer, autoformer, fedformer}. However, the scalability of these transformers on large datasets for LTSF tasks remains an open question \cite{kaplan2020scaling}. A recent study 
argues that the permutation-invariant nature of self-attention causes the loss of temporal information \cite{zeng2022transformers}. As a result, transformers often underperform compared to convolution-based models, potentially due to their struggles with local features and multi-scale features \cite{ts2vec, omniscale}. Overall, existing research on time-series transformers often lacks rigorous evaluation on large datasets and does not consistently outperform conventional approaches on small data. 

In this study, we provide an in-depth analysis of existing time-series Transformer models, covering key aspects such as the attention mechanism and position embedding. We argue that the seeming inadequacy of current transformer-based models for time-series data is due to their inability to model large-scale time series. Once these challenges are resolved, we would observe the  typical scaling law found in NLP and CV domains \cite{kaplan2020scaling, cv_scaling}. Motivated by this insight, we present a novel framework called \textbf{Timely Generative Pre-trained Transformer (\model)} (Fig. \ref{intro_figure}) that utilizes an extrapolatable position (xPos) embedding to encode trend and periodic patterns into time-series representations \cite{xPos}. \model integrates recurrent attention (also known as Retention) and convolution modules for effectively capturing both global temporal dependencies and nuanced local interactions \cite{retnet, conformer}. 

The key contributions of our research are threefold:
\begin{enumerate}
    \item We employ extrapolatable xPos embedding (Fig. \ref{intro_figure}b) to encode both trend and periodic patterns into time-series representations, facilitating long-term forecasting.
    
    \item We extend recurrent attention (Fig. \ref{intro_figure}c) to handle both continuous and irregularly-sampled time series data;
    
    \item We introduce convolution subsampling tokenizer (Fig. \ref{intro_figure}a) to extract features from raw time-series and temporal convolution (Fig. \ref{intro_figure}d) to sift local features among the timesteps.
\end{enumerate}
Overall, our experimental results reveal that \model effectively extrapolates temporal representations for long-term forecasting. This leads to highly effective pre-training on large-scale time-series biosignals and longitudinal EHR data, and ultimately superior task-specific fine-tuning performance compared to the existing methods.


\section{Related work}
\label{revisiting}

\subsection{Self-attention in Transformer}

Transformer employs an encoder-decoder architecture composed of $L$ layers of Transformer blocks \cite{transformer}. Each block consists of a self-attention layer followed by a feed-forward layer. For an input embedding $\mX \in \R^{N \times d}$, where $N$ is the number of tokens and $d$ is the hidden size, the self-attention mechanism is defined as:
\begin{equation}
\label{eq: attention}
\text{Attention}(\mX) = \text{Softmax}\left( \frac{\mQ \mK^\top}{\sqrt{d}} \right) \mV
\end{equation}
where $\mQ, \mK, \mV = \mX \mW_Q, \mX \mW_K, \mX \mW_V \in \R^{N \times d}$ are the Query, Key, and Value matrices, respectively. The attention mechanism allows Transformer to model long-term dependencies effectively, making it extensively utilized in NLP and CV domains. 

As one of the prominent time-series transformers, Conformer utilizes the self-attention mechanism to capture long-range global contexts in speech data \cite{conformer}. When combined with convolution modules, Conformer enhances self-attention by exploiting fine-grained local patterns. Although widely successful, the quadratic complexity of self-attention with respect to sequence length has spurred the exploration of attention-free modules such as Multi-Layer Perceptron (MLP) \cite{mlpmixer}, implicit long convolution \cite{hyena}, and Recurrent Neural Network (RNN) \cite{rwkv,retnet}. In particular, RNN-based attention modules have scaled up to 14 billion parameters while maintaining competitive performance with linear training and constant inference complexities. These modules are particularly well-suited for time-series modeling by effectively capturing sequential dependencies \cite{S4}. In this study, \model integrates the Retention mechanism and convolution modules to effectively capture both global and local contexts. 

\subsection{Position embedding in Transformer}

Transformer relies on position embedding to capture temporal relations, since the self-attention mechanism alone does not inherently discern token order \cite{relative_pe}. \emph{Absolute} position embedding, which commonly employs sinusoidal functions, adds positional encoding directly to token embeddings. 
However, this method only encodes discrete position indexes, making it less effective  for continuous timescales such as trend and periodic patterns in time-series data \cite{zeng2022transformers}. In contrast, speech transformers utilize \emph{relative} position embedding to handle continuous time by encoding positional information relative to token distances \cite{conformer}. Rotary Position Embedding (RoPE), prevalent in numerous large language models \cite{GPT3, touvron2023llama, penedo2023refinedweb}, applies rotation matrices to encode time information from relative distances \cite{rope}. Additionally, the RNN-based Transformer Receptance Weighted Key Value (RWKV) uses exponential decay to encode time information based on relative distance \cite{rwkv}. Bridging these techniques, xPos embedding utilizes both rotation and exponential decay to effectively capture long-term dependencies \cite{xPos}. 

One challenge for Transformer is \emph{extrapolation}, i.e., forecasting sequences longer than those seen during training, due to the difficulty in generalizing position embeddings to unseen positions \cite{alibi}. Encoder-decoder architectures often concatenate the input sequence with a zero-padded placeholder for the target sequence and predict all timesteps at once, while encoder-only models encode input sequence for forecasting \cite{informer, patchTST}. Both approaches struggle with extrapolation and rely heavily on their linear layer for forecasting \cite{li2023revisiting}, limiting their effectiveness in LTSF tasks. To address the issue, Attention with Linear Biases (ALiBi) adjusts attention with penalties linearly correlated with token distances \cite{alibi}. Building on this, xPos embedding employs exponential decay to assign penalties based on relative distances \cite{xPos}. Consequently, xPos can handle inference lengths up to eight times the training length while maintaining comparable performance. Our \model extends xPos from the NLP domain to  long-term forecasting in the time-series domain, focusing on exploring the underlying mechanisms that enable the temporal extrapolation.

\section{TimelyGPT Methodology}
Our proposed \model effectively pre-trains on unlabeled data using next-token prediction task to learn temporal representations (Fig. \ref{intro_figure}). It first processes time-series inputs using a convolution-subsampling tokenizer for token embedding (Fig. \ref{intro_figure}a). To extract meaningful temporal patterns, \model integrates three technical contributions. First, \model utilizes extrapolatable xPos embedding to encode trend and periodic patterns (Fig. \ref{intro_figure}b, Section \ref{sec:xPos}). Second, \model utilizes the Retention module to capture global content (Fig. \ref{intro_figure}c, Section \ref{sec: Retention}). Third, \model deploys the convolution module to capture the local content (Fig. \ref{intro_figure}d, Section \ref{sec: convolution}). Integrating Retention and Convolution modules enables the modeling of interactions between global and local content.

\subsection{Extrapolatable position embedding encodes temporal patterns }\label{sec:xPos}

As our first contribution,  \model employs xPos to encode relative positional information into token embeddings based on the distance $n-m$ between token $n$ and $m$ \cite{xPos}. Given an input embedding $\mX \in \R^{N \times d}$ for $N$ tokens at $d$ embedding dimensions, xPos is integrated into the $n$-th token embedding $\mX_n$ through rotation matrix $e^{i \theta n}$ and exponential decay $\gamma^n$:
\begin{align}\label{def:xPos}
    & \tilde \mQ_n  \tilde \mK_m = \mX_n \mW_Q (\gamma e^{i\theta})^{n-m} \mX_m \mW_K= \gamma^{n-m} \hat \mQ_n \hat \mK_m\nonumber \\
    &\text{where}\quad \hat \mQ_n = \mX_n \mW_{Q} e^{i\theta n}, \,  \hat \mK_m = \mX_m \mW_{K} e^{-i\theta m} 
\end{align} 
where $\theta$ and $\gamma$ indicate position-dependent rotation and decay hyperparameters \cite{rope, xPos}. The exponential decay $\gamma^{n-m}$ determines the intensity of remembering historical information, while the rotation matrix $e^{i\theta n}$ captures the oscillation frequencies. This decay mechanism effectively attenuates the influence of distant tokens, aiding in capturing long-term dependencies and enhancing extrapolation ability \cite{xPos}.

While initially designed for language modeling, xPos provides a compelling way for time-series modeling, mirroring the seasonal-trend decomposition (Fig. \ref{intro_figure}c). Its exponential decay $\gamma^{n-m}$ naturally concentrates on recent times while diminishing the influence of distant times, reflecting the trend momentum of time-series. The rotation matrix $e^{i \theta (n-m)}$ captures the seasonal component of  time-series through sinusoidal oscillations.

In healthcare time-series, xPos embedding effectively encodes both trend and periodic patterns crucial for   modeling continuous biosignals and irregular clinical records. For continuous biosignals, trend patterns such as body temperature and vital signs are key health indicators, while electrocardiograms (ECGs) exhibit periodic patterns reflecting the physiological rhythms of the human body. In irregularly-sampled clinical records, age-related susceptibility to illnesses is observed in longitudinal population studies using administrative health data \cite{ahuja2022mixehr, mixehr_seed}. Some EHRs also exhibit periodic patterns, especially for chronic diseases like COPD, which have alternating exacerbation and recovery cycles. 

We hypothesize that xPos embedding can encode these trend and periodic patterns into token embeddings. By harnessing xPos, \model can effectively model long-term dependencies essential for time-series forecasting. In Section \ref{sec: forecasting_exp}, \ref{sec: cls_ists_exp}, and \ref{sec:ablation}, we validated our hypothesis and explored the underlying mechanisms driving temporal extrapolation for forecasting beyond training length.

\subsection{Retention for continuous and irregularly-sampled time series} \label{sec: Retention}

As our second contribution,  we adapt the Retention mechanism to effectively handle continuous time-series data \cite{retnet}. The Retention mechanism based on xPos can be reformulated as an RNN to naturally model time-series data. Given the xPos embedding in Eq \ref{def:xPos}, the forward-pass of the Retention mechanism can be computed in parallel over all tokens with a linear training complexity: 
\begin{align}
\label{eq:Retention}
& \hat \mQ_n = \mX_n \mW_{Q} e^{i\theta n}, \:  \hat \mK_m = \mX_m \mW_{K} e^{-i\theta m}, \: \mV = \mX \mW_V 
\nonumber \\
& \text{Ret}(\mX) = (\hat \mQ \hat \mK^\top \odot \mD) \mV, \: \mD_{nm} = \begin{cases} \gamma^{n-m},   & n \geq m\\ 0, & n < m \end{cases}
\end{align}
where the decay matrix $\mD \in R^{N \times N}$ and rotation matrix $e^{i\theta (n-m)}$ encode trend and periodic patterns into token embedding, taking into account the distance between tokens $n-m$. When reformulated as an RNN, the Retention in Eq.~\ref{eq:Retention} can be manifested in a recurrent forward-pass with a constant inference complexity. This reformulated RNN excels in capturing sequential dependencies from the time-series. To handle long sequences, we use chunk-wise Retention by segmenting the sequence into multiple, non-overlapping chunks (Fig. \ref{intro_figure}c). Consequently, chunk-wise Retention maintains a linear complexity for long sequences. We provide details about the three Retention forward-passes in Appendix \ref{equivalence}. 

To accommodate irregularly-sampled time series, we modify the Retention mechanism as follows. Given $N$ samples $\{s_1, \ldots, s_N\}$, each sample $s_n$ is represented as a tuple $(x_n, t_n)$, consisting of an observation $x_n$ and a timestep $t_n$. 
Given two samples $s_n$ and $s_m$, the decay mask $\mD$ is adapted according to the time gap $\Delta t_{n, m} = t_n - t_m$:
\begin{equation} \label{eq: parl_ists}
    \text{Ret}(\mX) = (\mQ \mK^\top \odot \mD) \mV, \; 
    \mD_{nm} = 
    \begin{cases} 
        \gamma^{\Delta t_{n,m}},  & t_{n} \geq t_{m} \\ 
        0, & t_{n} < t_{m} 
    \end{cases}
\end{equation}

\begin{figure}[!h]
\centering
\includegraphics[width=\linewidth]{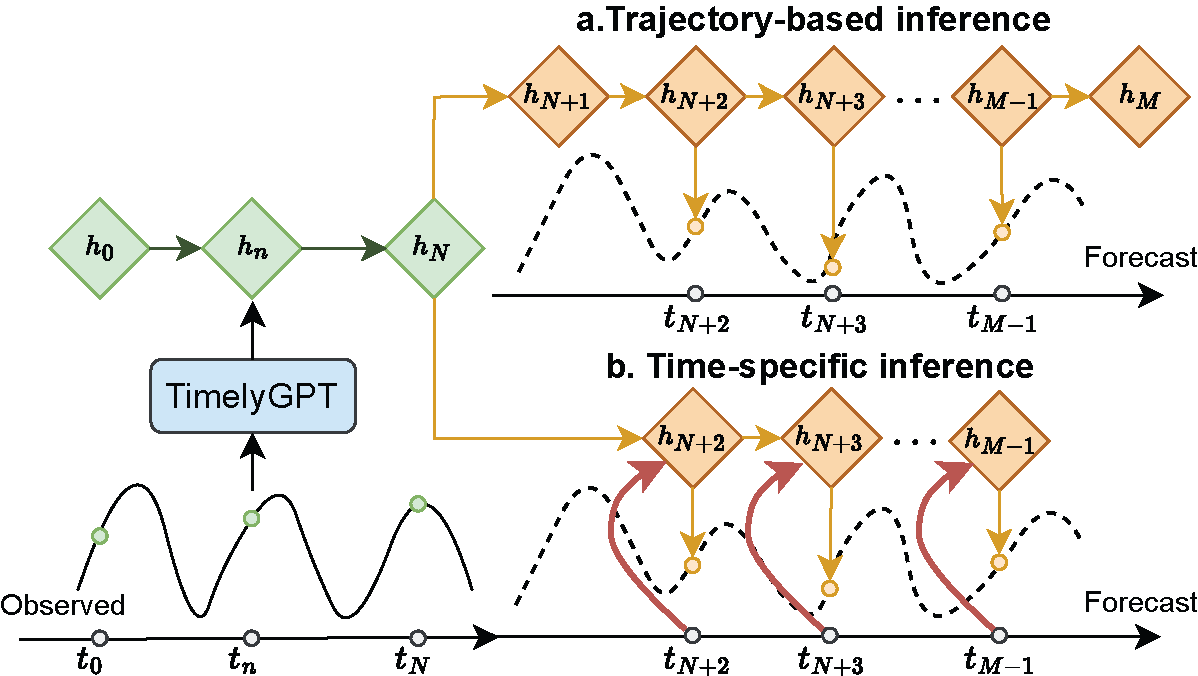}
\caption{Two inference strategies for forecasting irregularly-sampled time series. (a) Trajectory-based inference. \model autoregressively predicts the entire sequence at equal time intervals. The target intervals can then be taken from part of the inferred trajectory. (b) Time-specific inference. \model directly predicts the target data point using historical hidden states and the gap between the target timestep and the last observed timestep.}
\label{fig:inference_strategies}
\end{figure}
For next token pre-training, the retention incorporates $\Delta t_{n,n-1}$ into the recurrent state variable $\mS_n \in R^{d \times d}$, :
\begin{align} \label{eq: recur_ists}
    & \mS_{n} = \gamma^{\Delta{t}_{n,n-1}} \mS_{n-1} + \mK_{n}^\top \mV_{n} \nonumber \\
    &\text{Ret}(X_{n}) = \mQ_{n} \mS_{n} 
\end{align}
where the base case $\mS_1=\mathbf{0}$ in this recurrent relation.


At inference time, to forecast irregularly-sampled time series, we consider two recurrent inference strategies, namely trajectory-based inference and time-specific inference (Fig. \ref{fig:inference_strategies}). Both strategies make predictions based on a look-up window. The former autoregressively predicts a trajectory at equal time intervals. The latter directly makes prediction at a specific time point $s_{n'} = (x_{n'}, t_{n'})$. Specifically, knowing the target timestep $t_{n'}$ and the last observed sample $s_{n} = (x_{n}, t_{n})$, \model outputs the embedding of the target token $\text{Ret}(X_{n'})=\mQ_{n'}\mS_{n'}$, taking into account the time gap $\Delta t_{n', n} = t_{n'} - t_{n}$ and the recurrent state then becomes $\mS_{n'} = \gamma^{\Delta{t}_{n',n}} \mS_{n} + \mK_{n}^\top \mV_{n}$.


\subsection{Convolution modules for local interaction} \label{sec: convolution}
Convolution methods excel at identifying localized interactions from time series \cite{convolution}. As the first part of our third contribution, we propose a \textbf{convolution-subsampling tokenizer} for feature extraction from the raw time-series input (Fig. \ref{intro_figure}a). Briefly, it uses multiple 1-D convolution layers to condense the time dimension and extract local features of the time-series. The convolution-subsampling tokenizer consists of two 1-D convolution layers with kernel size 3 and stride 2, reducing the sequence length to 1/4. Unlike the prevalent patching technique, which merely segments adjacent timesteps and features \cite{patchTST}, the convolution tokenizer effectively captures local temporal interactions. More details are provided in Appendix \ref{sec: overflow}.

As the second part of our third contribution, we propose a \textbf{temporal convolution module} using a depth-wise separable convolution \cite{depth_sep_conv}, sifting local temporal features from the time-series representations. As shown in Fig. \ref{intro_figure}d, this module starts with a layer normalization, followed by a 1-D depth-wise convolution and a point-wise convolution layer, with batch normalization and swish activation after the depth-wise convolution. Integrating convolution and attention allows \model to extract global-local feature interactions \cite{wu2020lite, conformer}. By stacking multiple decoder layers,  each with a convolution module, \model discerns multi-scale features that characterize patterns across varying time scales \cite{omniscale}.

\subsection{Computational complexity}

\model with its efficient Retention mechanism achieves $O(N)$ training complexity and $O(1)$ inference complexity. In contrast, BERT and GPT  incur $O(N^2)$ training complexity and $O(N)$ inference complexity \cite{linear_attention}. The vanilla attention mechanism in the Transformer, $\text{Attention}(X) =\text{Softmax} (\frac{QK^T}{\sqrt{d}}) V$, introduces a training complexity of $O(N^2d)$. This quadratic computational bottleneck prevents standard Transformer models from modeling long sequences (i.e., $N >> d$). 

\model achieves linear training complexity by following research in linear transformers \cite{linear_attention}. In the Retention mechanism, $\text{Ret}(X_n) =Q_n S_n, S_n=K_n^T V_n + \gamma S_{n-1}$, both $Q_n S_n$ and $K_n^T V_n$ have $O(d^2)$ complexity. By recursively updating over $N$ timesteps, the total complexity becomes $O(N d^2)$. For inference, \model proposes time-specific and trajectory-based methods. The trajectory-based inference recursively generates sequences with equally-spaced time intervals like the GPT model, incurring $O(N)$ inference complexity. In contrast, the time-specific inference directly predicts target time point with $O(1)$ complexity.
Therefore, \model achieves $O(N)$ training complexity and $O(1)$ inference complexity, making it computationally efficient and suitable for long sequences. We provided detailed discussion of computational bottleneck of Transformer and efficient linear Transformer in Appendix ~\ref{sec: linear_attention_review}.

\section{Data}\label{sec:data}

\subsection{Sleep-EDF dataset}
\label{sec:biosignal_data}
The Sleep European Data Format (EDF) database, sourced from PhysioBank \cite{physiobank}, contains sleep recordings from 153 healthy subjects \cite{sleep}.  These whole-night polysomnographic sleep recordings include 7 types of biosignalS: electroencephalogram (EEG) from Fpz-Cz and Pz-Oz electrode locations, electrooculogram (EOG), submental chin electromyogram (EMG), oro-nasal airflow, rectal body temperature, and an event marker. Both EEG and EOG signals were sampled at 100 Hz (i.e., the signals were recorded at a rate of 100 samples per second), while EMG and the other features were sampled at 1 Hz (i.e., 1 sample per second). 
Sleep patterns (hypnograms) were manually scored by trained technicians into five sleep stages. 
This biosignal dataset comprises a total of 1.2 billion timesteps, segmented into 300,700 sequences of 4,000 timesteps each. It provides large-scale continuous time-series data for training large models. In our experiment, we forecast all 7 biosignals.

\subsection{PopHR database} \label{sec:PopHR_data}
The Population Health Record (PopHR) database hosts a massive amount of longitudinal claim data from the provincial government health insurer in Quebec, Canada (Régie de l’assurance maladie du Québec, RAMQ) on health service use \cite{pophr1, pophr2}. In total, there are approximately 1.3 million participants in the PopHR database, which represents a randomly sampled 25\% of the population in the metropolitan area of Montreal between 1998 and 2014. Cohort memberships are maintained dynamically by removing deceased residents and actively enrolling newborns and immigrants. We extracted irregularly-sampled time series from the patient clinical records in the PopHR database. Specifically, we converted ICD-9 diagnostic codes to phenotype codes (PheCodes) using the expert-defined \href{https://phewascatalog.org/phecodes}{PheWAS catalog} \cite{phewas1, phewas2}.
we selected 315 unique PheCodes each with over 50,000 token counts and excluded patients who had fewer than 50 PheCode tokens. This resulted in a dataset of 489,000 patients, averaging 112 diagnosis records each.

\section{Experiments}

We first validated the scaling pattern of \model, determining the optimal number of model parameters for different dataset sizes (Section \ref{sec: scaling_exp}). We then explored \model's extrapolation capabilities for long-term forecasting up to 6,000 timesteps in Sleep-EDF's biosignal data, and analyzed extrapolation's underlying mechanism through visualization (Section \ref{sec: forecasting_exp}). Our evaluation extended forecasting to irregularly-sampled time series (Section \ref{sec: cls_ists_exp}). Furthermore, we conducted ablation studies to evaluate the contributions of various components (Section \ref{sec:ablation}).

\subsection{Pre-training and fine-tuning}

During pre-training, \model utilizes a next-token prediction task to learn general temporal representations from unlabeled data \cite{GPT2}. Given a sequence with a [SOS] token, \model predicts the subsequent tokens by shifting the sequence to the right. At the last layer, each token's output representation is fed into a linear layer for next-token prediction. The pre-training loss is Mean Squared Error (MSE) for continuous signals (e.g., biosignal) and cross-entropy for discrete signals (e.g., diagnosis codes). 

Among other Transformer baselines, PatchTST adopted a masking-based approach, masking 40\% of its patches as zeros \cite{patchTST}. CRT utilized a dropping-based pre-training, discarding up to 70\% of patches \cite{CRT}. For the Transformer models without established pre-training methods, we used a masking-based method by randomly masking 40\% of timesteps \cite{TST}.For downstream forecasting tasks, we employ end-to-end fine-tuning on the entire model. The final linear layer is utilized for making the forecasts. All Transformer models performed 20 epochs of pre-training with MSE loss, followed by 5 epochs of end-to-end fine-tuning.

\subsection{Jointly forecasting multivariate  biosignals from Sleep-EDF dataset}

We utilized all seven features from the Sleep-EDF dataset for a multivariate forecasting task, applying standardization as preprocessing.
The Sleep-EDF dataset was split into training (80\%), validation (10\%), and test (10\%) sets. All models were pre-trained on the entire training set and fine-tuned on a randomly chosen 20\% subset of the training data, with time-series data segmented into non-overlapping sequences. For pre-training, we chose an input length of 4,000 timesteps. For fine-tuning, we used a look-up window of 2,000 timesteps and varied forecasting windows of 720, 2,000, and 6,000 timesteps. We used MAE as a metric. We evaluated \model against Informer \cite{informer}, Autoformer \cite{autoformer}, FEDformer \cite{fedformer}, PatchTST \cite{patchTST}, TimesNet \cite{wu2023timesnet}, TS2Vec \cite{ts2vec}, and DLinear \cite{zeng2022transformers}. Based on the scaling law in Section \ref{sec: scaling_exp}, we set the model parameters for all transformers to around 18 million, with specific architectures and parameters detailed in Table~\ref{tab:timelyGPT_baseline_setup}.


\subsection{Forecasting irregularly-sampled diagnostic codes from PopHR dataset}

We assessed long-term forecasting task of the  irregularly-sampled time series extracted from the PopHR database. We divided the dataset into training (80\%), validation (10\%), and testing (10\%) sets. We pre-trained on the entire training set and fine-tuned on a 20\% subset of training data. We used cross entropy and top-$K$ recall to evaluate the pre-training and fine-tuning, respectively. For forecasting, we set the look-up window to be 50 timestamps and the rest as the forecasting window, containing up to more than 100 timestamps (i.e., diagnosis codes). 

For our \model, we separately evaluated the performance of trajectory-based and time-specific inferences (Section \ref{sec: Retention}). We compared with several transformer baselines, including Informer, Fedformer, AutoFormer, and PatchTST as well as the models designed for irregularly-sampled time series, namely mTAND \cite{shukla2021multitime} and SeFT \cite{SeFT}. 
Given that diagnoses are discrete values, there was no need to utilize the convolution-subsampling tokenizer for \model. Furthermore, we specified a patch size of 2 for PatchTST, indicating that every two adjacent timestamps are projected into a single patch. Based on the scaling law in Section \ref{sec: scaling_exp}, we set model parameters for all transformers to about 7.5 million, with specific architectures and parameters detailed in Table~\ref{tab:timelyGPT_baseline_setup}.  

\subsection{Model parameters}
\begin{figure}[b]
\centering
\includegraphics[width=\columnwidth]{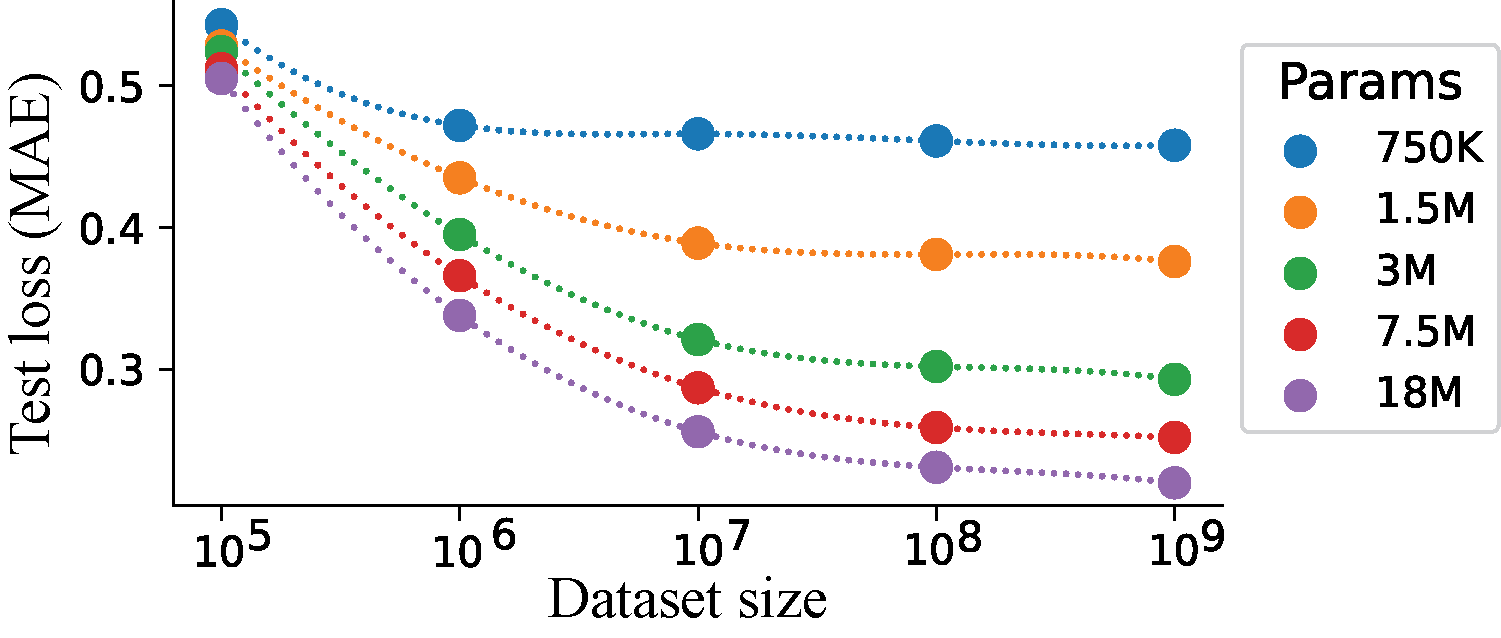}
\caption{Test MAE of forecasting Sleep-EDF biosignals as a function of dataset sizes and parameter sizes. Both look-up and forecasting windows were set to 256 timesteps. \model with more parameters tends to exhibit better performance when trained on larger datasets.}
\label{scaling_law}
\end{figure}

For all benchmark experiments, we tailored the architecture and parameters of \model based on the scaling-law analysis (Section \ref{sec: scaling_exp}; Fig.~\ref{scaling_law}). Specifically, for the Sleep-EDF dataset, \model was configured with 18 million parameters, and for the PopHR dataset, it was configured with 7.5 million parameters. While different Transformer models may have unique optimal hyperparameters, optimizing each model's setup is computationally prohibitive with our current compute resources. For fairness of comparison, we compared \model against all transformer baselines at the same model size (Table~\ref{tab:timelyGPT_baseline_setup}).

\begin{figure*}[t]
\centering
\includegraphics[width=\textwidth]{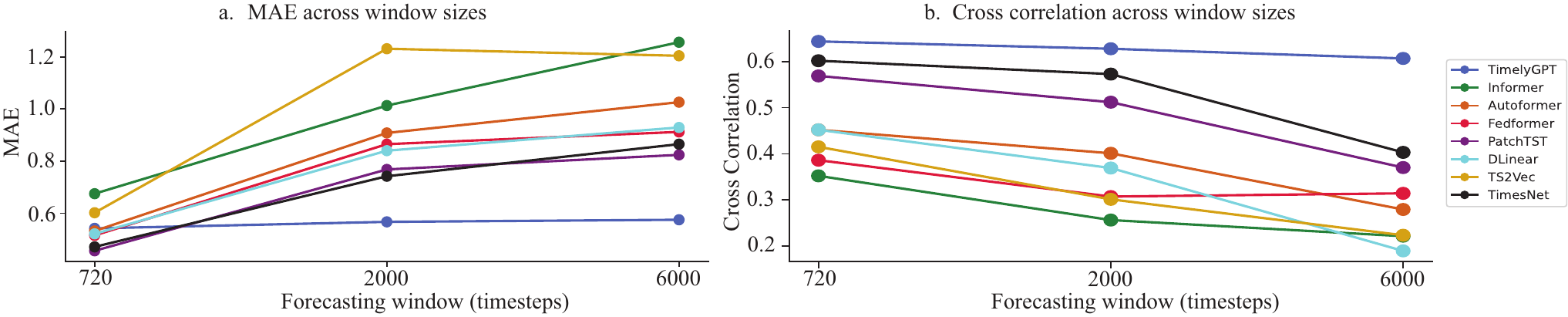}
\caption{SleepEDF biosignal forecasting performances of \model and seven state-of-the-art methods over various forecasting windows. \textbf{a.} MAE for 8 methods evaluated over 3 forecasting windows (720, 2000, and 6000 timesteps). \textbf{b.} Cross-correlation scores for the same methods and forecasting windows. The detailed numerical results are summarized in Table~\ref{tab:forecasting_comp}.}
\label{forecasting_cls_exp_quantitative}
\end{figure*}

\section{Results}

\subsection{Scalability of \model} 
\label{sec: scaling_exp}

We evaluated the scalability of \model on the large-scale Sleep-EDF dataset to determine the optimal model parameters with respect to different dataset sizes \cite{sleep}. We selected subsets of the Sleep-EDF dataset with timesteps ranging from $10^5$ to $10^9$, splitting each dataset into training (80\%), validation (10\%), and testing (10\%) sets. Both look-up and forecasting windows were set to 256 timesteps for this experiment. \model's performance improves as parameter and dataset size increase (Fig. \ref{scaling_law}), which is attributed to its capacity to handle more data, known as the scaling law for Transformer \cite{kaplan2020scaling}. 
We provide further discussion of scaling patterns of the existing Transformer models in Appendix ~\ref{sec: scaling_law}.

\begin{table}[b]
\caption{Comparison of \model as well as 7 baselines for long-term forecasting experiment on the large-scale SleepEDF dataset. Bold and underlined numbers indicate the best and second best results for each metric and window.}
\label{tab:forecasting_comp}
\begin{center}
\begin{tabular}{l|c|c|c|c|c|c}
\hline
Window Size & \multicolumn{3}{c|}{MAE} & \multicolumn{3}{c}{Cross-Correlation} \\
\hline
Metrics & 720 & 2000 & 6000 & 720 & 2000 & 6000 \\
\hline
\model  & 0.542 & \textbf{0.567} & \textbf{0.575} & \textbf{0.644} & \textbf{0.628} & \textbf{0.607} \\
Informer   & 0.675 & 1.013 & 1.256 & 0.352 & 0.256 & 0.221 \\
Autoformer & 0.532 & 0.908 & 1.026 & 0.452 & 0.401 & 0.279 \\
Fedformer  & 0.515 & 0.865 & 0.912 & 0.386 & 0.307 & 0.314 \\
PatchTST   & \textbf{0.456} & 0.768 & \underline{0.824} & 0.569 & 0.512 & 0.370 \\
DLinear    & 0.521 & 0.840 & 0.929 & 0.452 & 0.369 & 0.189 \\
TS2Vec     & 0.602 & 1.231 & 1.204 & 0.415 & 0.301 & 0.223 \\
TimesNet   & \underline{0.471} & \underline{0.742} & 0.865 & \underline{0.602} & \underline{0.573} & \underline{0.403} \\
\hline
\end{tabular}
\end{center}
\end{table}

\begin{figure*}[t!]
\centering\includegraphics[width=0.75\textwidth]{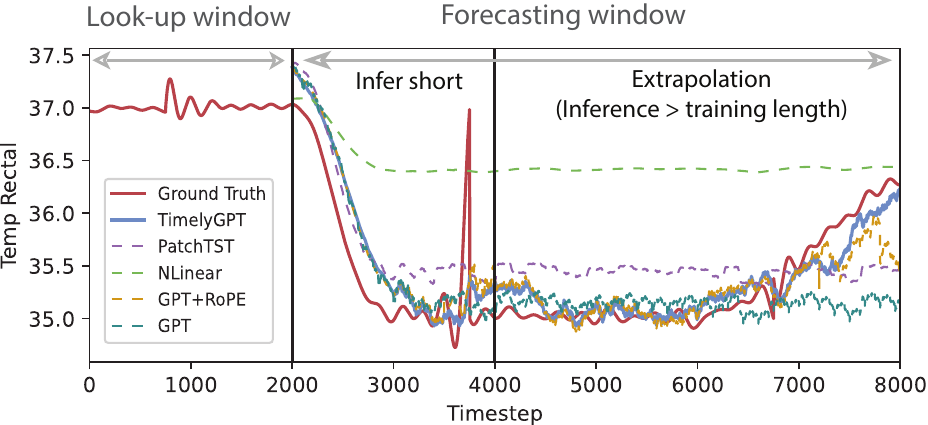}
\caption{Predicted sequence of SleepEDF biosignals of 6,000 timesteps. Given a 2,000 look-up window, we applied \model (blue solid line) and 4 state-of-the-art methods (dashed lines) to predict the biosignals for the next 6,000 timesteps. The groundtruth biosignals are displayed as red solid line. The two vertical lines demarcate the look-up window and the length of pre-training sequences, respectively.
}
\label{forecasting_cls_exp}
\end{figure*}

\subsection{Forecasting multivariate Sleep-EDF biosignals} 
\label{sec: forecasting_exp}
\model achieved the best performance in forecasting biosignals for all windows in terms of MAE, except for window 720 (Table \ref{tab:forecasting_comp}; Fig. \ref{forecasting_cls_exp_quantitative}a).
PatchTST achieved the best MAE at 0.456, whereas \model conferred comparable performance. DLinear was also effective for the 720-timestep forecasting window. However, as the forecasting window increased to 2,000 and 6,000 timesteps, both PatchTST and DLinear suffered performance drops due to their reliance on the linear layers and inability to extrapolate beyond the training length. In contrast, pre-trained on 4,000 timesteps,  \model consistently maintained superior performance up to 6,000 timesteps given a short look-up window (i.e., prompt) containing only 2,000 timesteps. Additionally, \model consistently outperformed other baselines across all three forecasting windows in terms of cross-correlation performance (Fig. \ref{forecasting_cls_exp_quantitative}b; Table \ref{tab:forecasting_comp}). TimesNet was the second best performer for these windows, but declined as window size gets larger due to the extrapolation issue. These results underscore \model's extrapolation capabilities in long-term forecasting, aligning with the findings in the NLP domain \cite{xPos}.



We visualized the predicted biosignals by \model against the leading baselines (PatchTST and DLinear) and the ablated methods (GPT-2 and GPT-2 with RoPE), focusing on sleep stage transitions (Fig. \ref{forecasting_cls_exp}). We utilized a 2,000-timestep look-up window and a 6,000-timestep forecasting window. Forecasting beyond 2,000 timesteps is marked as extrapolation, as it exceeds the training length. In the rectal temperature (i.e., trend signal), \model's forecast aligned well with the groundtruth, effectively capturing distinct trend patterns. Notably, the small bump in the prompt before the 1000-th timestep is a typical indicator for temperature drop. Most models were able to capture it except for DLinear, showing the benefits of pre-training. Beyond the training length of 4000, \model demonstrated more advantages in accurately extrapolating the rise of the rectal temperature around 7000-th timestep while PatchTST and GPT fell behind. The superior extrapolation capabilities of \model is attributable to its ability to capture the long-term trends with xPos embedding. In contrast, both PatchTST and vanilla GPT experienced a performance decline, likely due to the dependency on linear mapping as discussed in previous research \cite{li2023revisiting}. Additionally, \model exhibits superior extrapolation capabilities over the ablated baseline GPT+RoPE, highlighting its effective trend pattern modeling for extrapolation. We also visualized EEG periodic biosignal forecast and found a similar conclusion (Fig. \ref{fig:visualization_period}).

\subsection{Forecasting patient diagnosis trajectory}
\label{sec: cls_ists_exp}

\begin{figure}[b]
    \centering
    \includegraphics[width=\linewidth]{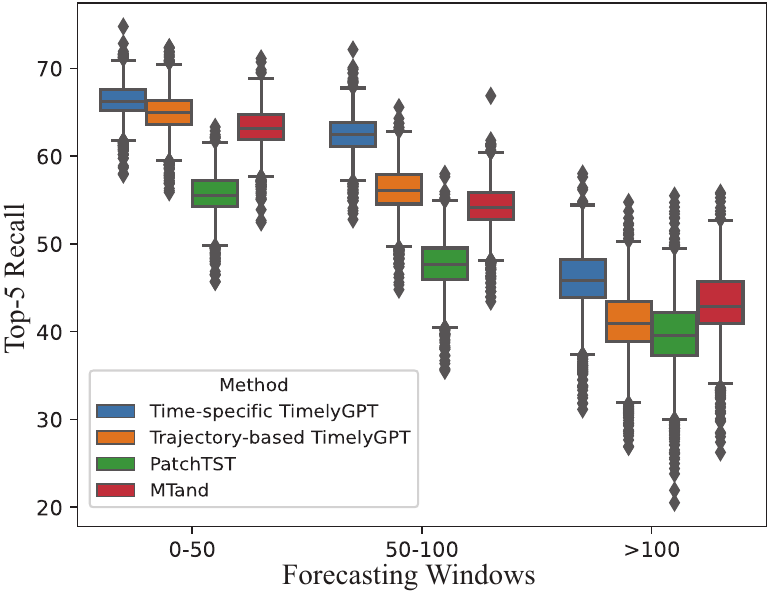}
    \caption{The distribution of top-5 recall performance for \model with two inference methods (Time-specific and Trajectory-based), compared to PatchTST and MTand across three forecasting window sizes.}
    \label{fig:topk_distribution}
\end{figure}

\begin{figure*}[!t]
    \centering
    \includegraphics[width=\textwidth]{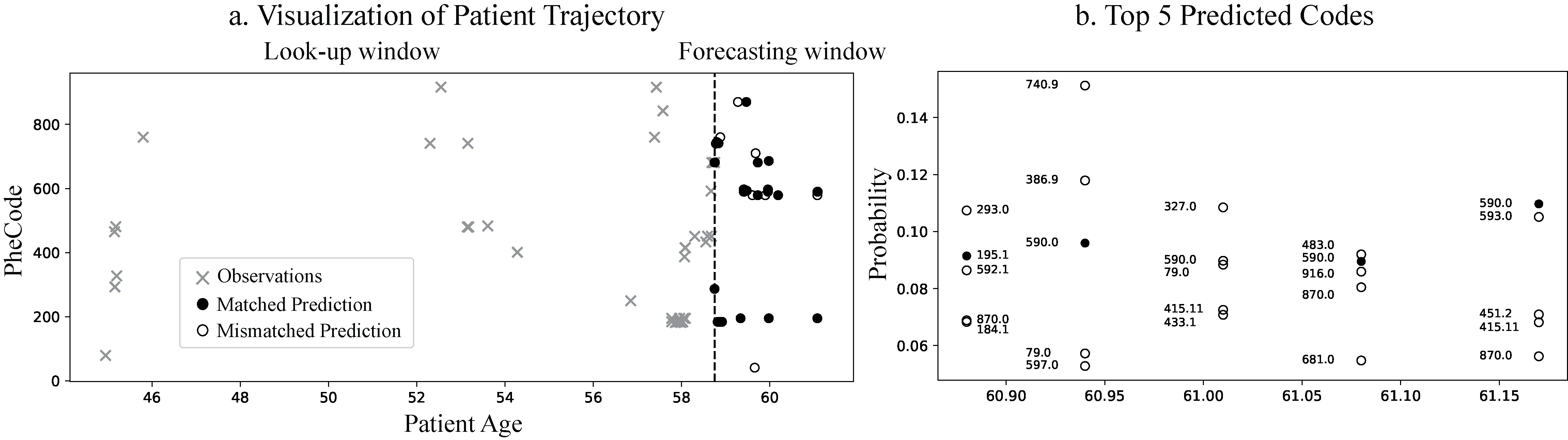}
    \caption{Visualization of a cancer patient's medical trajectory from the PopHR dataset. \textbf{a.} Look-up and forecast windows. Matched predictions (solid circles) were identified when the top 5 predicted PheCodes contain the groundtruth. \textbf{b.} The top 5 predicted PheCodes for the final 5 timesteps of the subject.}
    \label{fig:vis_pophr}
\end{figure*}

We then applied \model and the baseline methods to forecast 315 PheCodes for 489K patients from PopHR (Section \ref{sec:PopHR_data}). We evaluated the performance using the average top K recall at each forecast window. \model with time-specific inference outperformed the baselines reaching the highest recall rates of 58.65\% and 70.83\% at $K=5$ and $K=10$, respectively  (Table~\ref{tab:forecast_ists}). At $K=15$, \model ranked the second-highest recall with 82.69\%. In addition, the time-specific inference outperformed the trajectory-based inference, highlighting the advantage of time decay mechanism. 


\begin{table}[t]
\caption{Forecasting results of \model and 6 baselines on PopHR's irregular-sampled time series dataset. \model with time-specific inference achieved the highest recall at $K=5$ and $K=10$, and the second highest at $K=15$, demonstrating its superior performance in long-term forecasting of irregularly-sampled time series.  }
\label{tab:forecast_ists}
\centering
\begin{tabular}{lccc}
\toprule
\multirow{2}{*}{\textbf{Metrics}} & \multicolumn{3}{c}{\textbf{Recall @\(K\) (\%)}} \\
 & \(K = 5\) & \(K = 10\) & \(K = 15\) \\
\midrule
\textbf{\model (trajectory-based)} & 52.30 & 64.35 & 77.12 \\
\textbf{\model (time-specific)} & \textbf{58.65} & \textbf{70.83} & \underline{82.69} \\
\hline
Informer & 46.37 & 60.14 & 71.24 \\
Autoformer  & 42.87 & 57.43 & 68.59 \\
Fedformer & 43.31 & 58.34 & 69.60 \\
PatchTST & 48.17 & 65.55 & 73.31 \\
MTand & \underline{52.59} & \underline{70.21} & \textbf{83.73} \\
SeFT & 49.26 & 68.10 & 79.39 \\
\bottomrule
\end{tabular}
\end{table}

We then examined the distributions of the top-5 recall rates at 3 forecast windows, comparing two inference methods of \model with the best transformer baseline PatchTST and the leading irregular time series algorithm MTand (Fig. \ref{fig:topk_distribution}). \model's time-specific inference consistently outperformed trajectory-based inference as the forecasting window size increases. While both inference methods exhibited similar performance for predicting the first 50 timesteps, time-specific \model demonstrated significantly better results beyond 50 timesteps. This improvement is likely due to time-specific inference taking into account the evolving states and the query timestep in the time decay mechanism, enhancing its ability to predict the temporal evolution of healthcare trajectories over irregular intervals. As expected, all models  experienced a performance decline in predicting farther future because of the increasing uncertainties. Despite this, \model maintained higher and more stable performance within the first 100 steps compared to PatchTST and MTand. Although MTand closely followed to time-specific \model for the first 50 timesteps, its performance drastically declines as the forecasting window increases, reflecting its difficulty with extrapolation. These findings highlight the utility of the proposed time-specific inference in leveraging time-decay mechanism to handle irregularly-sampled time series for long-term forecasting.

We visualized the observed and predicted trajectory of a patient with neoplasm and genitourinary diseases (Fig. \ref{fig:vis_pophr}). \model with time-specific inference produced a high top-5 recall rate of 85.7\% on this patient. Indeed, most of the observed codes were among the top 5 predicted codes by the time-specific \model. Zooming into the forecast window (Fig. \ref{fig:vis_pophr}b), \model accurately predicted Phecodes 590.0 (Pyelonephritis) three times around the age of 61. \model predicted PheCode 740.9 at age 61 with high probability, which appeared twice at ages 52 and 53 in the look-up window. Therefore, \model demonstrated a promising direction to forecast patient health state despite the challenges inherent in modeling irregularly-sampled longitudinal EHR data.

\begin{table}[b]
\centering
\caption{Ablation results of \model w/o specific components, showing forecasting performance for a 6,000-timestep window in the Sleep-EDF dataset and top-15 recall rate in the PopHR dataset.}
\label{tab:ablation_table}
\begin{tabular}{l|cc}  
\toprule
\multirow{2}{*}{\textbf{Datasets}} & \textbf{Sleep-EDF} & \textbf{PopHR} \\
& (6000) & ($K$=5) \\
\midrule
\model (with Pre-training) & \textbf{0.575} & \textbf{58.65} \\
\; \; w/o Convolution Subsampling & 0.587 & --- \\
\; \; \; \; w/o Temporal Convolution & 0.581 & 57.69 \\
\; \; \; \; \; \; w/o Exponential Decay & 0.715 & 52.50 \\
\; \; \; \; \; \; \; \; w/o RoPE (GPT-2) & 1.072 & 50.18 \\
\hline
\model (w/o Pre-training) & 0.641 & 56.42 \\
\bottomrule
\end{tabular}
\end{table}

\subsection{Ablation study}
\label{sec:ablation}

To assess the contributions of various components in \model, we conducted ablation studies by omitting the key components, including convolution subsampling tokenizer, temporal convolution module, exponential decay, and RoPE relative position embedding. Notably, removing all components results in a vanilla GPT-2. Since exponential decay in xPos depends on RoPE, we cannot assess the impact of exponential decay independently by removing the RoPE component. Additionally, we also ablated the pre-training strategy by training \model from scratch on the forecasting tasks. The ablation studies focused on downstream forecasting experiments using the Sleep-EDF and PopHR datasets, corresponding to continuous biosignals and irregularly-sampled time series, respectively. We conducted the ablation on long-term forecasting of 6000 timesteps in the Sleep-EDF dataset and evaluated the top-5 recall scores in the PopHR dataset.

As shown in Table~\ref{tab:ablation_table}, for the Sleep-EDF forecasting task, removing the RoPE component led to the most significant performance degradation (a MAE of 0.357).
The removal of exponential decay also led to increase MAE of 0.134, demonstrating its benefits of encoding trend patterns for long-term forecasting. Together, the two ablation experiments show the importance of xPos as our first main contribution (Section \ref{sec:xPos}). 
The integration of convolution modules helps \model capture local features, although the benefits were smaller compared with other components. 

In the forecasting of irregularly-sampled time series, the exponential decay and RoPE components improved performance by 6.15\% and 2.32\%, respectively. The time decay mechanism encodes trend patterns into the modeling of patients' health trajectories, making it a promising approach for forecasting irregular clinical diagnoses. Pre-training decreased MAE by 0.066 for forecasting  continuous biosignals in Sleep-EDF and increased top K recall rate by 2.21\% for forecasting irregularly sampled diagnostic codes. 

\section{Conclusion and Future Work}

\model effectively forecasts long sequences of time-series, utilizing xPos embedding, recurrent attention, and convolution modules. For continuously monitored biosignals such as Sleep-EDF, \model can accurately extrapolate up to 6,000 timesteps given only a 2000-timestep prompt. Moreover, \model also effectively forecasts irregularly-sampled time series by conditioning the recurrent Retention on the time. 
In our future work, we will perform comprehensive and in-depth analysis on the trajectory inference of the EHR data, as it may have a profound impact on the future of patient care and early intervention. \model is a causal model with unidirectional attention \cite{alibi}. This may limit its expressiveness in terms of time-series representation learning, which may be improved via a bidirectional architecture. To enhance transfer learning, we will adapt \model for out-of-distribution biosignals, further enhancing its utility in healthcare time-series.   


\bibliographystyle{ACM-Reference-Format}
\bibliography{sample-base}

\clearpage
\onecolumn
\appendix

\setcounter{table}{0}
\renewcommand{\thetable}{S\arabic{table}}%
\setcounter{figure}{0}
\renewcommand{\thefigure}{S\arabic{figure}}
\setcounter{page}{1}%

\section{Revisiting Transformers} \label{related_works}

\subsection{Efficient attention in Transformer}
\label{sec: linear_attention_review}
Transformer models have found extensive applications in both the Natural Language Processing and Computer Vision domains \cite{transformer}.  In the vanilla self-attention mechanism, the query, key, value matrices are denoted as $\mQ, \mK, \mV \in \R^{N \times d}$. The output embedding for the $n$-th token is represented as $\mO_n = \frac{\sum_m^N \text{sim}(\mQ_n, \mK_m) \mV_m}{\sum_m^N \text{sim}(\mQ_n, \mK_m)}$, where the the similarity function represents the softmax of inner-product  $\text{sim}(\mQ_n, \mK_m) = \exp( \mQ_n \mK_m^\top / \sqrt{d})$. The self-attention mechanism, also known as token-mixer, aims to integrate information from every token and thus capture global-range interaction. However, computing the dot product $\mQ_n \mK_m^\top$ before the softmax operation introduces computational complexity of $O(N^2 d)$. As sequence length increases, this quadratic complexity becomes bottleneck, making it challenging to train for longer sequences. Many studies have been proposed to address the quadratic issue in self-attention mechanism. The linear attention replaces the softmax term $\text{sim}(\mQ_m, \mK_m)$ with $\phi(\mQ_n)\phi(\mK_m^\top)$ for a nonlinear kernel function $\phi(\cdot)$ \cite{linear_attention}, avoiding quadratic computation.

Recent research has explored alternatives to the token-mixer attention mechanism including Multi-Layer Perceptron (MLP) \cite{mlpmixer}, convolution \cite{hyena}, and RNN \cite{rwkv,retnet}. Particularly, RNN-variant models like RWKV and RetNet have successfully scaled up to more than 14 billion parameters, yielding comparable performance to conventional transformers. A fascinating connection between linear attention and RNNs has been identified  \cite{linear_attention}, making RNN-based token mixer as efficient as linear attention. The output embedding from linear attention can be recast as an RNN: $\mO_n = \frac{\phi(\mQ_n) \sum_m^N\phi(\mK_m^\top) \mV_m}{\phi(\mQ_n) \sum_m^N \phi(\mK_m^\top)} = \frac{\phi(\mQ_n) \mS_n}{\phi(\mQ_n) \mZ_n}$, where $\mS_n = \sum_m^N\phi(\mK_m^\top) \mV_m, \, \mZ_n = \sum_m^N \phi(\mK_m^\top)$. Thus, the output embedding $\mO_n$ depends on both $\mS_n$ and $\mZ_n$, which are incrementally updated through cumulative sums. Thus, the RNN-based token-mixer not only competes in performance, but also offers linear training and consistent inference complexities. By employing exponential decay mechanism, it diminishes the influence of distant positions, transitioning from ``token-mixing" to ``time-mixing". Considering RNN's historical effectiveness in time-series and audio domains, it stands out as an excellent choice for temporal modeling.

\subsection{Time-series Transformer} 

Transformers are increasingly applied in LTSF tasks, attributed to their capabilities in capturing long-term temporal dependencies \cite{informer, autoformer, fedformer, etsformer, patchTST}. Researchers have modified transformers by incorporating custom attention modules to address complex temporal dependencies \cite{informer, autoformer, fedformer}. Studies like \cite{autoformer, fedformer, etsformer}  have introduced time-decomposition techniques into attention mechanisms to bolster modeling capability. The majority of studies focus on the encoder-decoder architecture, coupled with a one-forward prediction framework \cite{informer}. In this design, the decoder takes a concatenated input of the context (or prompt) and placeholder forecasting windows, directly generating the resulting embedding without autoregressive decoding. As a result, these models aim to avoid error accumulation seen in autoregressive frameworks, but aligning its performance closely with linear models \cite{zeng2022transformers}. Encoder-only models, like patchTST, use the encoded embedding for forecasting with the help of a linear layer \cite{patchTST}. Additionally, self-supervised representation learning techniques in time series, such as TS2Vec and TimesNet, offer valuable representation learning capabilities for forecasting tasks \cite{ts2vec, wu2023timesnet}.



\subsection{Transformer scaling law in time-series} \label{sec: scaling_law}

Despite the broad applications of transformer-based models in time-series data such as speech \cite{conformer, whisper}, biosignals \cite{EEG}, and traffic flow \cite{STEP, pdformer}, their effectiveness in capturing temporal dependencies in LTSF task has been limited and often underperforms compared to linear models \cite{zeng2022transformers}. As Table~\ref{transformer_setup} indicates, time-series transformer models often have much more parameters than the dataset size (timestep) with only two exceptions, namely large-size Conformer and CRT. Such disparities imply that many  transformers may be over-parameterized, leading to highly variable performance. In Section \ref{sec: scaling_exp}, our study validates the Transformer scaling law in time-series domain (i.e., scaling up both model parameters and dataset size to improve performance) \cite{kaplan2020scaling, cv_scaling}. For all benchmark experiments,  our proposed \model~effectively pre-trains on large-scale data with model parameters aligned to this scaling law.

\begin{table*}[b!]
\caption{The model parameters and utilized datasets of time-series transformers and comparison methods. These setups are sourced from papers and default implementation. Over-parameterization indicates model parameters $>>$ dataset size (timestep). 
}
\label{transformer_setup}
\begin{center}
\begin{tabular}{lcccccccc}
\toprule
\multicolumn{1}{c}{\bf Method} & Application & Dimension & Layer & Model Parameter & Dataset Size (Timestep) &  Param versus Data \\
\midrule 
Informer & Forecasting & 512 & 3 & 11.3M  &  69.7K & Over-param \\
Autoformer & Forecasting & 512 & 3 & 10.5M  &  69.7K & Over-param \\
Fedformer (F/W) & Forecasting & 512 & 3 &  16.3/114.3M & 69.7K & Over-param \\
PatchTST & Forecasting & 128 & 3 & 1.2M & 69.7K & Over-param \\
DLinear & Forecasting & - & 1 & 70K &  69.7K &  Adequate \\
Conformer (L) & Classification & 512 & 18 & 118.8M & 55.9B & Adequate \\
CRT & Pre-training &  128 & 18 &  8.8 M &  109.2M & Adequate \\
\bottomrule
\end{tabular}
\end{center}
\end{table*}

\begin{figure*}[t]
\centering
\includegraphics[width=\textwidth]{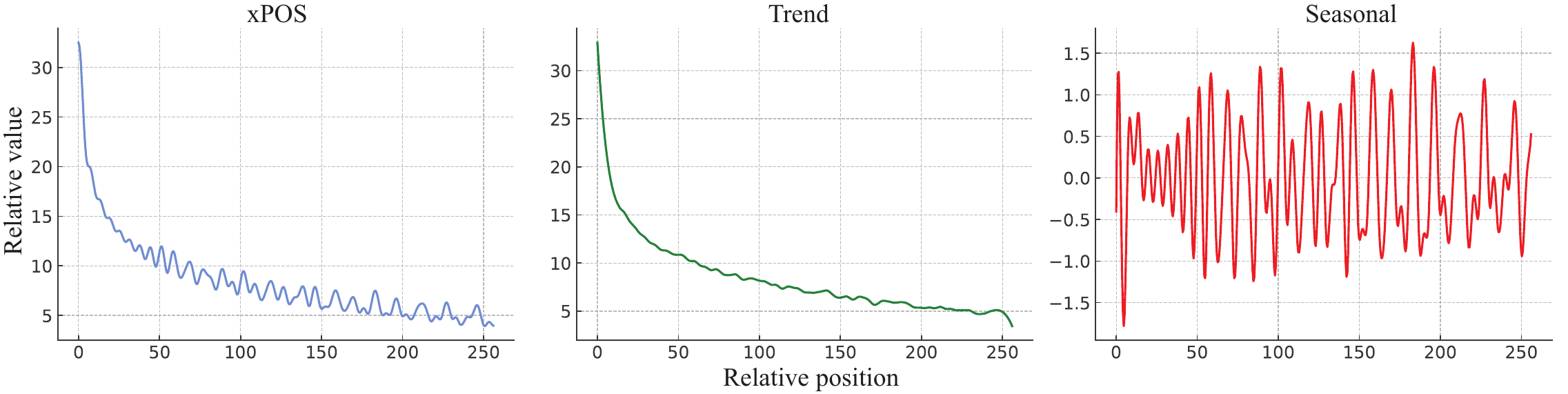}
\caption{The xPos embedding diminishes distant temporal information according to the relative distance, enabling decomposition to capture both trend and periodic dynamics in time-series data. 
}
\label{xPos}
\end{figure*}

\section{Details about TimelyGPT}

\subsection{From absolute to relative position embedding} \label{pos_emb}

Unlike RNNs or CNNs, the inclusion of positional embedding is essential for the Transformer model. Since the permutation-invariant self-attention mechanism cannot capture input order, making it challenging to differentiate tokens in various positions. The solution fall into two categories: (1) incorporate position information into the inputs, i.e., absolute position embedding; (2) modify the attention matrix to distinguish tokens at different positions, referring to relative position embedding. 

In absolute position embedding, the token representation for a given token $n$ consists of a word embedding $\mX_n$ and a position embedding $\mP_n$. The self-attention mechanism is expressed as:
\begin{align}
    & \mQ_n = (\mX_n + \mP_n) \mW_Q, \quad \mK_n = (\mX_n + \mP_n) \mW_K, \quad \mV_n = (\mX_n + \mP_n) \mW_V \nonumber \\
    & \mA_{n,m} = \text{softmax} (\mQ_n \mK_m^\top), \quad \mO_m = \sum_m \mA_{n,m} \mV_m
\end{align} 
where $\mA_{n,m}$ is an attention score between token $n$ and $m$ without scaling. The inner-dot product $\mQ_n \mK_m^\top$ and output embedding $\mO_m$ can be expanded as follows:

\begin{align} \label{qk_complete}
    &\mQ_n \mK_m^\top = (\mX_n + \mP_n) \mW_Q ((\mX_m + \mP_m) \mW_K)^\top \nonumber\\
    & = (\mX_n + \mP_n) \mW_Q \mW_K^\top (\mX_m + \mP_m)^\top \nonumber\\
    & = \underbrace{\mX_n \mW_Q \mW_K^\top \mX_m^\top}_{\text{token-token}} + \underbrace{\mX_n \mW_Q \mW_K^\top \mP_m^\top}_{\text{token-position}} +  \underbrace{\mP_n \mW_Q \mW_K^\top \mX_m^\top}_{\text{position-token} } + \underbrace{\mP_n \mW_Q  \mW_K^\top \mP_m^\top }_{\text{position-position}} 
\end{align}
\begin{equation}
    \mO_n = \sum_m \text{softmax} \left((\mX_n \mW_Q + \mP_n \mW_Q) (\mW_K^\top \mX_m^\top + \mW_K^\top \mP_m^\top)\right) (\mX_n + \mP_m) \mW_V 
\end{equation} 
where attention arises from four types of interactions: (1) token-token interaction; (2) token-position interaction; (3) position-token interaction; (4) position-position interaction. However, absolute position embedding only incorporates fixed position information, neglecting the relative positional difference between the token $n$ and $m$.





In the realm of audio processing, prevalent transformers like Conformer \cite{conformer} incorporate relative positional information through the T5 position embedding \cite{t5}. Notably, the T5 model suggests a minimal interaction between tokens and positions, resulting in the exclusion of token-position and position-token terms from the attention matrix:
\begin{align} \label{t5}
    \mQ_n \mK_m^\top &= \mX_n \mW_Q \mW_K^\top \mX_m^\top + {\color{red} \beta_{n,m}}
\end{align}
where the position-position interaction term, $\mP_n \mW_Q  \mW_K^\top \mP_m^\top$, is replaced with a trainable bias related to the position $n$ and $m$. The T5 position embedding follows Transformer-XL, omitting the position term $\mP_m \mW_V$ in the attentive aggregation computation \cite{transformerxl, xlnet}. As a result, the relative position embedding is only added to the dot product $\mQ \mK^\top$:

\begin{equation}
     \mO_n =\sum_m \text{softmax} (\mX_n \mW_Q \mW_K^\top \mX_m^\top + {\color{red} \beta_{n,m}} ) \mX_m \mW_V
\end{equation}
The RoPE technique leverages the property of rotation matrix to model positional information \cite{rope}. To incorporate this relative position information into the queries $\mQ$ and keys $\mK$, the method aims to identify functions $f_{\mQ}(\mQ, \cdot)$ and $f_K(\mK, \cdot)$ that satisfies this invariant criteria about relative distance:
\begin{equation} \label{B:relative}
    \left \langle \mQ_n, \mK_m \right \rangle = \left \langle  f_{\mQ}(\mQ, n), f_K(\mK, m) \right \rangle = g(\mQ, \mK, m-n),    
\end{equation} 
where $g$ is a function that depends only on the relative distance $m-n$ and $\mQ = \mX \mW_Q$ and $\mK = \mX \mW_K$ stand for token embedding for queries and keys matrices, respectively. RoPE defines the function $f$ involving a $d$-dimensional rotation matrix $\mR$: 
\begin{equation} \label{B:rope}
    f_{\mQ}(\mQ, n) = \mR^{d}_{\Theta, n} (\mX_n \mW_Q) , \quad f_{\mK}(\mK, m) = \mR^{d}_{\Theta, m} (\mX_m \mW_K) 
\end{equation}   
With a given hidden size $d$, a block diagonal matrix $\mR^{d}_{\Theta, n}$ contains multiple rotation matrices $(\mR^{(1)}_{n, \theta_{1}}, \dots, \mR^{(d/2)}_{n, \theta_{d/2}})$ on its diagonal:
\begin{align} \label{B:rope_R}
\mR^{d}_{\Theta, n} = \begin{bmatrix}
\mR^{(1)}_{n, \theta_1}  &        &  \\
                        & \ddots &  \\
                        &        & \mR^{(d/2)}_{n, \theta_{d/2}}
\end{bmatrix}, \quad
\mR^{(i)}_{n, \theta_{i}} = \begin{bmatrix}
\text{cos} \ n \theta_{i} & -\text{sin} \ n \theta_{i} \\
\text{sin} \ n \theta_{i} & \text{cos} \ n \theta_{i}
\end{bmatrix}
\end{align}
where the rotation hyperparameter $\theta_i = 10000^{-2(i-1)/d}$. In RoPE, any even-dimension representation can be built by placing multiple 2-dimensional rotation matrices diagonally within the $\mR^{d}_{\Theta, n}$ matrix, expanding hidden size from 2-dimension to $d$-dimension. As $\mR^{d}_{\Theta, m-n} = (\mR^{d}_{\Theta, n})^\top \mR^{d}_{\Theta, m}$, RoPE satisfies the property outlined in Eq \ref{B:relative}:

\begin{align}
& \left \langle \mQ_n, \mK_m \right \rangle = \sum_{i=1}^{d/2} \left \langle \mQ_n[2i-1:2i], \mK_m[2i-1:2i] \right \rangle \nonumber \\
&= \sum_{i=1}^{d/2} \mR^{d}_{\theta_i, m-n} \left \langle (\mX_n \mW_Q) [2i-1:2i], (\mX_m \mW_K) [2i-1:2i] \right \rangle
\end{align}
In RoPE, relative position information is added to the inner product $\mQ \mK^\top$ by rotating the angles of queries and keys matrices. Recently, \cite{xPos} argues that the sinusoids used in the rotation matrices do not change monotonically. Instead, they oscillate dramatically as the relative distance increases. This limitation hinders RoPE's ability to sequences of extended lengths. To address it, \cite{xPos} proposes xPos  that preserves the advantage of ROPE and behaves stably at long-term dependency by measuring position monotonicity \cite{xPos}.

\subsection{Equivalence of three forward-pass Retention}\label{equivalence}
According to Section \ref{sec: Retention}, the parallel forward-pass is equivalent to the recurrent forward-pass. With the initial state variable $\mS_0 = 0$, the recurrent forward-pass can be expressed as follows:
\begin{align}
    & \textbf{Recurrent: } \mS_n  = \underbrace{\mK_n^\top \mV_n }_{\text{Single-token}} + \gamma \mS_{n-1} ,\quad \text{Ret} (\mX_n) = \mQ_n \mS_n \nonumber \\
    & \implies \mS_n = \sum_{m}^n \gamma^{n-m} \mK_m^\top \mV_m ,\quad \text{Ret} (\mX_n) = \mQ_n \sum_{m}^n \gamma^{n-m} \mK_m^\top \mV_m
\end{align}
where $\text{Ret} (\mX_n)$ calculates the Retention at single-time $n$ by considering timestep $i$ up to the current time. It corresponds to the $n$-th timestep (row) of parallel forward-pass of Retention.
\begin{align}
    & \textbf{Recurrent: } \text{Ret} (\mX_n) = \mQ_n \sum_{m}^n \gamma^{n-m} \mK_m^\top \mV_m  \nonumber \\
    & \implies \textbf{Parallel: } \text{Ret} (\mX_n) =  \underbrace{\mQ_n}_{1 \times d_{qk}} 
    \underbrace{\mK_{m\leq n}^\top}_{d_{qk} \times n}
    \underbrace{\odot \mD_{m\leq n} }_{n \times n}
    \underbrace{\mV_{m\leq n}}_{n \times d_v}
\end{align}
When the recurrent forward-pass traverses all timesteps, the parallel and recurrent forward-passes of Retention become identical. With the parallel and recurrent forward-passes of Retention, we aim to show the equivalence between the chunk-wise forward-pass and the parallel and recurrent forward-passes. The computation of chunk-wise Retention involves both parallel intra-chunk and recurrent inter-chunk computation as follows. 

\begin{align}
    & \textbf{Chunk-wise: } \text{Ret}(\mX_{[i]}) = \underbrace{(\mQ_{[i]} \mK_{[i]}^\top \odot \mD) \mV_{[i]} }_{\text{Intra-chunk}}  + \underbrace{(\mQ_{[i]} \mS_{[i-1]}) \odot \zeta}_{\text{Inter-chunk}}   \nonumber \\
   & \mS_{[i]} = \underbrace{\mK_{[i]}^\top (\mV_{[i]} \odot \mD_B)}_{\text{Current chunk}}+ \underbrace{\gamma^B \mS_{[i-1]} }_{\text{Past chunk}}, \, \zeta_{j} =\gamma^{j}
\end{align}
where $\zeta = [\gamma^1, \gamma^2, \ldots, \gamma^B]^\top$ is a column-vector of time-decay scaling factor for inter-chunk attention between the current chunk $[i]$ and the previous chunk $[i-1]$. Specifically, $\gamma^j$ is the scaling factor for the $j^{th}$ row of chunk $[i]$ from the last row of chunk $[i-1]$ such that the bigger the $j$ index the smaller the $\gamma^j$ value.
Therefore, Retention recursively aggregates information from the $i$-th chunk (i.e., intra-chunk embedding) and the previous chunk (i.e., inter-chunk embedding). 


For the per-chunk state variable $\mS_{[i]}$, it computes current-chunk information as well as past-chunk information. The current-chunk information $\mK_{[i]}^\top \mV_{[i]}$ decays by $\mD_B$, which is the last row of decay matrix $\mD$. The past chunk information $\mS_{[i-1]}$ is decayed with respect to the chunk size $B$. The initial state variable $\mS_{[i=0]} = 0$ is computed recurrently given the chunk size $B$:
\begin{equation}
    \mS_{[i]}  = \mK_{[i]}^\top (\mV_{[i]} \odot \mD_{B}) + \gamma^B \mS_{[i-1]} = \sum_{m=1}^B \gamma^{B-m} \mK_m^\top \mV_m + \gamma^B \mS_{[i-1]}
\end{equation}

Moreover, the update of state variable $\mS_{[i]}$ can be reformulated in parallel. The first term represents the information of current chunk, and the second term represented the past-chunk information decayed by the chunk size $\mB$. Consequently, $\mS_{[i-1]}$ represents the state information from the beginning to the $(i-1)$-th chunk, and we represent the inter-chunk information in chunk-wise Retention:
\begin{align}
    & \mS_{[i-1]} = \sum_{m=1}^{B*i} \gamma^{B*i-m} \mK_m^\top \mV_m = \mK^\top_{1:(B*i)} \odot \mD_{1:(B*i)} \mV_{1:(B*i)} \nonumber \\
    & \underbrace{(\mQ_{[i]} \mS_{[i-1]}) \odot \zeta}_{\text{Inter-chunk}} = (\mQ_{(B*i):(B*(i+1))}  \mK^\top_{1:(B*i)}  \odot \mD_{1:(B*i)} \mV_{1:(B*i)}) \odot \zeta \nonumber \\
    & = \mQ_{(B*i) : (B*(i+1))} \mK^\top_{1:B*i} \odot \mD_{(B*i) : (B*(i+1))} \mV_{1:(B*i)}
\end{align}
where the intra-chunk computation updates each row of the lower triangular matrix (highlighted as green in Fig. \ref{intro_figure}.c). Together, the recurrent intra-chunk computation with the parallel intra-chunk computation (highlighted as purple Fig. \ref{intro_figure}c) completes the chunk-wise forward-pass of Retention. 
\begin{figure*}[h]
\begin{center}
\includegraphics[width=\textwidth]
{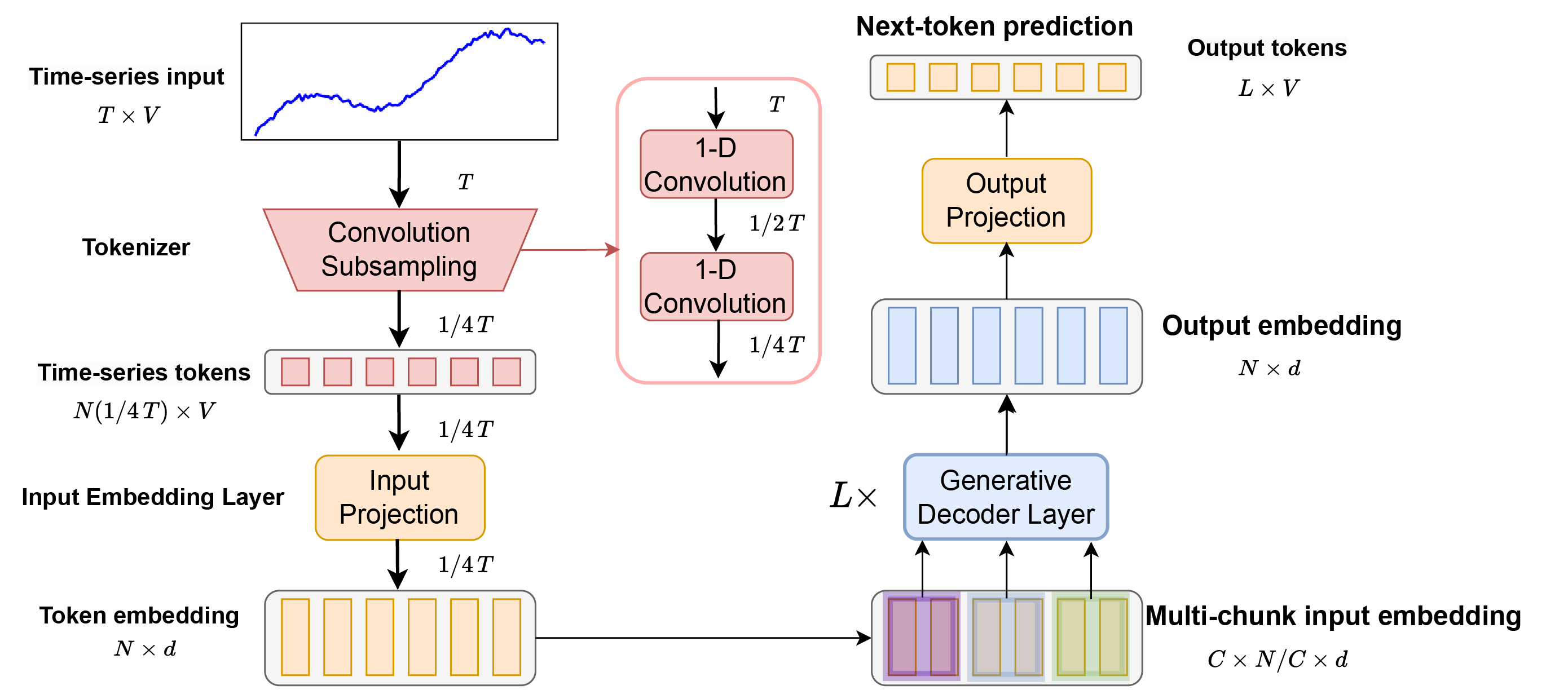}
\end{center}
\caption{Schematic of the TimelyGPT Pre-Training Process}
\label{scheme_pretraining}
\end{figure*}

\subsection{TimelyGPT pre-training overflow}
\label{sec: overflow}

For the \model pre-training, we illustrate the full processes of input processing, model training, and next-token prediction in Fig. \ref{scheme_pretraining}. For a time-series input with $T$ timesteps and $V$ variates, it is tokenized via a convolution-subsampling module. This tokenizer, typically comprising two 1-D convolution layers with a kernel size of 3 and stride of 2. It produces a sequence of tokens of the shape $N \times V$, effectively reducing the  sequence length to 1/4, i.e., $N= 1/4T$. The sequence of tokens is projected into an input embedding of the shape $L \times d$ with an linear projection layer. As a result, the input embedding is passed through $L$ generative decoder layers, where the Retention mechanism takes segmented mulitple-chunk input embedding. Finally, the output embedding of the shape $N \times d$ is passed through an output projection layer, which generate a sequence of tokens with the shape of $L \times V$ for next-token prediction.

\section{Experiment Summary}
\label{sec: app_exp_results}

\begin{table*}[h]
\caption{Configurations of \model , transformer baselines, and recurrent models across different   datasets}
\label{tab:timelyGPT_baseline_setup}
\begin{center}
\setlength\tabcolsep{4pt} 
\begin{tabular}{lcccccc}
\toprule
& \bf Sleep-EDF & \bf PopHR \\
\midrule
Data Size (timesteps) & 1.2B & 54.9M \\
Model Parameters & 18M & 7.5M \\
\hline
\multicolumn{3}{l}{\bf \model} \\
\hline
Decoder Layers & 12 & 8 \\
Heads & 8 & 4 \\
Dim ($\mQ$, $\mK$, $\mV$, FF) & 320,320,640,640 & 200,200,400,400 \\
\hline
\multicolumn{3}{l}{\bf Transformer baselines including Encoder-decoder and Encoder-only models} \\
\hline
Enc-Dec Layers & 6 \& 6 & 4 \& 4 \\
Encoder Layers & 12 & 8 \\
Decoder Layers & 12 & 8 \\
Heads & 8 & 4 \\
Dim ($\mQ$, $\mK$, $\mV$, FF)  & 384,384,384,1536 & 200,200,200,400 \\
\hline
\multicolumn{3}{l}{\bf Recurrent Models} \\
\hline
Layers & 12 & 8 \\
Dim & 384 & 200 \\
\bottomrule
\end{tabular}
\end{center}
\end{table*}

\begin{figure*}[h]
    \centering
    \includegraphics[width=0.6\linewidth]{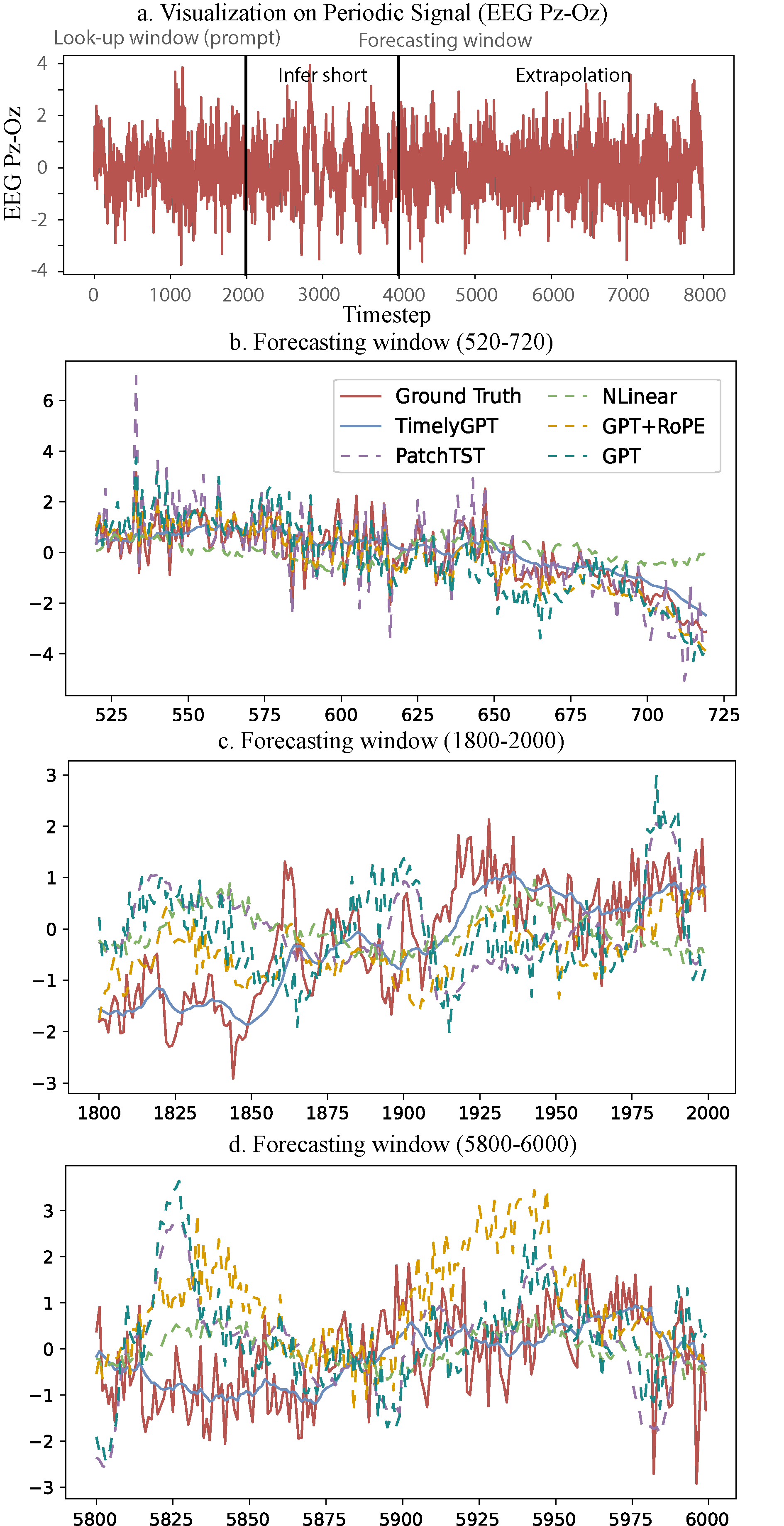}
    \caption{Example of forecasting experiments on the period signal (EEG Pz-Oz).  \textbf{a}. the groundtruth of EEG Pz-Oz singal. Forecasting results are shown between 520 and 720 timesteps (\textbf{b}), 1800 and 2000 timesteps (\textbf{c}), and 5800 and 6000 timesteps (\textbf{d}).  \model~is able to forecast the periodic signals up to 6000 timesteps owing to the extrapolation capabilities. }
    \label{fig:visualization_period}
\end{figure*}

We summarize the setup of model architecture for \model and other baselines for the experiments in Table~\ref{tab:timelyGPT_baseline_setup}. Additionally, we also provide the visualization of forecasting experiment on the period signal (EEG Pz-Oz) in Fig. \ref{fig:visualization_period}.


\end{document}